%% file: claf.tex
\documentclass{article}
\usepackage{spconf,amsmath,graphicx}

\usepackage{multirow}
\usepackage{algorithm,algorithmic}
\usepackage{amsfonts}
\usepackage{amssymb}
\usepackage{physics} 
\usepackage{dsfont} 
\usepackage{listings}

\usepackage{etoolbox}
\makeatletter
\AfterEndEnvironment{algorithm}{\let\@algcomment\relax}
\AtEndEnvironment{algorithm}{\kern2pt\hrule\relax\vskip3pt\@algcomment}
\let\@algcomment\relax
\newcommand\algcomment[1]{\def\@algcomment{\footnotesize#1}}
\renewcommand\fs@ruled{\def\@fs@cfont{\bfseries}\let\@fs@capt\floatc@ruled
  \def\@fs@pre{\hrule height.8pt depth0pt \kern2pt}%
  \def\@fs@post{}%
  \def\@fs@mid{\kern2pt\hrule\kern2pt}%
  \let\@fs@iftopcapt\iftrue}
\makeatother

\usepackage{varioref} % references to show page
\usepackage[pagebackref=true,breaklinks=true,letterpaper=true,colorlinks,bookmarks=false]{hyperref}
\usepackage[capitalise]{cleveref} 
\crefname{appsec}{appendix}{appendices}
\Crefname{appsec}{Appendix}{Appendices}

\title{Robustness through Cognitive Dissociation Mitigation in Contrastive Adversarial Training}
\name{Adir Rahamim, Itay Naeh}
\address{Rafael - Advanced Defense Systems Ltd., Israel\\
        adir.r6@gmail.com, itay@naeh.us}
\begin{document}
%\ninept
%
\maketitle
\begin{abstract}
In this paper, we introduce a novel neural network training framework that increases model's adversarial robustness to adversarial attacks while maintaining high clean accuracy by combining contrastive learning (CL) with adversarial training (AT). We propose to improve model robustness to adversarial attacks by learning feature representations that are consistent under both data augmentations and adversarial perturbations. We leverage contrastive learning to improve adversarial robustness by considering an adversarial example as another positive example, and aim to maximize the similarity between random augmentations of data samples and their adversarial example, while constantly updating the classification head in order to avoid a cognitive dissociation between the classification head and the embedding space. 
This dissociation is caused by the fact that CL updates the network up to the embedding space, while freezing the classification head which is used to generate new positive adversarial examples. 
We validate our method, \textit{Contrastive Learning with Adversarial Features(CLAF)}, on the CIFAR-10 dataset on which it outperforms both robust accuracy and clean accuracy over alternative supervised and self-supervised adversarial learning methods.
\end{abstract}
\begin{keywords}
Adversarial defense, adversarial training, contrastive learning
\end{keywords}

\input{section/intro}
\input{section/related_work}
\input{section/method}
\input{section/experiments}
\input{section/conclusion}
\input{section/ablation}
% -------------------------------------------------------------------------
\bibliographystyle{IEEEbib}
\bibliography{refs}

\end{document}

%% file: section/intro.tex
\section{Introduction}
\label{sec:intro}

In recent years, Deep Neural Networks (DNNs) have
shown phenomenal performance in a wide range of tasks, but at the same time, these networks are susceptible to adversarial attacks\cite{szegedy2013intriguing, goodfellow2014explaining, papernot2016limitations} -- malicious small (sometimes imperceptible) perturbation added to an image that alters the output of the network. The model's ability to resist adversarial attacks is critical for real-life applications, such as autonomous vehicles. Various methods have been proposed to achieve trained models with better robustness to adversarial attacks including  \cite{xie2017mitigating, madry2017towards, zhang2019theoretically}. Among them, the most promising is adversarial training: train the model using both clean and adversarial inputs generated by some attacker, e.g Fast Gradient Sign Method(FGSM)\cite{goodfellow2014explaining}, Projected Gradient Descent(PGD)\cite{madry2017towards} and TRADES\cite{zhang2019theoretically}.\\

\noindent Recently, Self-Supervised Learning (SSL) has emerged as a technique for learning representation without the need for labeled data. Representations obtained by self-supervised pretraining can be easily transferred to other downstream tasks with promising results, for example, discriminating between
image rotations\cite{gidaris2018unsupervised}. Contrastive learning proves to be an effective SSL technique with promising results\cite{he2020momentum, chen2020simple}. Specifically, CL frameworks are trained to learn the representation of unlabeled data by choosing an anchor and pulling it and its positive samples together in embedding space, while at the same time pushing it far apart from many negative samples.
Since CL framework assumes no labels are available, the positive pair is often different augmentations of the sample, and a negative pair often consist of the sample and randomly chosen samples from the current batch. Recent research, Supervised Contrastive Learning (SCL) \cite{khosla2020supervised} extends the CL framework by leveraging label information, where embeddings from the same class are pulled closer together than embeddings from different classes. This enriches the set of positive samples per anchor, by using samples of the same class as the anchor, in addition to many negative samples.\\

\noindent In this work, we propose to consider an adversarial example as a positive example. Now, given the clean sample data augmentations and its adversarial counterpart, we pull together their feature representation in the embedding space and push it apart from many negative samples, to achieve better clean and robust accuracy. We will adopt the recent SCL framework for better results.\\

\noindent A question that must be asked is how to craft the adversarial samples during contrastive learning, as adversarial images are images with intentionally perturbed pixels that cause \textbf{misclassification}, and in contrastive learning framework we only have an encoder and a projection head model and no linear classifier, which is necessary for the creation of adversarial samples. 
To overcome this hurdle, we leverage linear evaluation protocol\cite{chen2020simple}. Within its learning process, the network is practically divided into two separately trained sections: the encoder, from the input to the embedding space, which is trained contrastively, and the classification head, which is trained with labeled data. Producing an adversarial example requires the whole network. When training the encoder, it changes and dissociates itself from the classification head a bit every epoch, until the embedding space represents information in a way which is irrelevant to the classification head. This cognitive dissociation between the two parts causes the previously trained classification head to be unsuitable in developing new adversarial examples for the updated encoder.
We will need to maintain the linear classifier up-to-date with the current feature representations learned by the encoder, as in each encoder training epoch we learn new latent representation space, and using an outdated classifier will result in feature inconsistency and irrelevant adversarial examples. 
We suggest that for every encoder training epoch, we add a training, for a fixed and relatively small number of epochs, a linear classifier on top of the frozen encoder, that will be used to craft the adversarial samples that will be used in SCL encoder training. Note that the classification head is a single linear layer, thus its training is relatively fast, and it does not add substantial overhead to the total training time.\\

\noindent With this in mind, we introduce \textit{Contrastive Learning with Adversarial Features(CLAF)}, a supervised contrastive adversarial training framework to achieve state-of-the-art model robustness. CLAF consists of an encoder, projection head, and classifier. We start with a regular SCL phase for a fixed number of epochs where we train only the encoder and projection head to let them stabilize. Then, we add the adversarial samples to the set of positive samples. To do so, in each training step, we first train for a fixed and relatively small number of epochs a linear classifier, then we freeze it and use it on top of the encoder to create the desirable adversarial examples. In the last phase, we freeze the encoder parameters and train a final classification head.

%% file: section/related_work.tex
\section{Related Work}
\label{sec:related}

\noindent\textbf{Adversarial robustness.} Adversarial robustness is an emerging topic in deep learning since \cite{szegedy2013intriguing} first showed their existence. To increase model robustness, \cite{goodfellow2014explaining} proposed the adversarial training method. First, it generates adversarial inputs by some attacker, then these examples are used to update the network parameters. 
This process can be interpreted as solving the following min-max optimization problem:
\begin{equation} \label{eq:0}
\min_\theta \frac{1}{n} \sum_{i=1}^n \max_{x_i'\in \mathcal{B}(x_i,\epsilon)} \ell (f_\theta (x_i'), y_i)
\end{equation}
Where $\mathcal{D} = {(x_i,y_i)}_{i=1,...,n}$ denotes the train set composed of pairs of clean image $x_i$ with $y_i$ as its label,  $f_\theta(x)$ represents a model parameterized by weighs $\theta$. $\ell$ denotes the cross-entropy loss and $\mathcal{B}(x_i,\epsilon)$ denotes the $\ell_p$-norm ball centered at $x_i$ with radius $\epsilon$, where the most common norm is $\ell_\infty$. The inner maximization problem aims to find an adversarial version of the current sample $x_i$ that maximizes the training loss. 

\noindent Over the years, several methods proposed to generate adversarial examples. An early method by \cite{goodfellow2014explaining} proposed the fast gradient sign method(FGSM), which generates adversarial examples with a single gradient step that maximizes the loss.
Follow-up work by \cite{madry2017towards} using projected gradient descent(PGD) to create models with better robustness to multi-step attacks. PGD is an iterative variant of the gradient-based attack with the addition of multiple random restarts. 
Since then, the PGD attack and the adversarial training framework have been tested with different variations, for example, penalize the difference between logits of clean samples and their adversarial counterparts\cite{kannan2018adversarial}, use momentum to improve adversary\cite{dong2018boosting}, use of different activation function\cite{xie2020smooth} and generalization to multiple types of adversarial attacks\cite{tramer2019adversarial, maini2020adversarial}.

\noindent Recent TRADES\cite{zhang2019theoretically} method achieves high robustness levels by replacing the cross-entropy loss of adversarial training with a loss that promotes logits pair between the natural sample logits and the adversarial sample logits.\\

\noindent Beside adversarial training, other defense mechanisms have been proposed, for example, DefenseGAN\cite{samangouei2018defense} method trains a GAN to learn the data distribution to generate a denoised version of adversarial example. Perturbation Rectifying Network (PRN) \cite{akhtar2018defense} use 'pre-input' layers to the model, and if a perturbation is detected, we use the PRN output for label prediction instead of the actual image. feature squeezing\cite{xu2017feature} is a method proposed to reduce the search space available to an adversary by merging samples that correspond to many different feature vectors in the original space into a single sample.\\

\noindent\textbf{Self-supervised learning.} In recent years, self-supervised learning has gained great popularity in the computer vision community, because of the costly operation of creating labeled datasets. SSL is generally divided into two parts: Pretraining a model on a pretext task using unlabeled data only for learning a good feature representation, then fine-tuning it for downstream tasks(e.g., image classification) using (few-shot) labeled data. Extensive studies have been conducted on self-supervised learning \cite{gidaris2020learning, caron2018deep}. Among these, methods based on contrastive learning\cite{hadsell2006dimensionality, oord2018representation} achieved state-of-the-art results, with learned features that surpass the  feature learned in a supervised manner on many downstream tasks\cite{chen2020simple, he2020momentum}. CL feature representation learning is done by maximizing the agreement of positive pairs(two random augmentations of the original sample) relative to a large number of negative pairs. Recent work proposed SCL\cite{khosla2020supervised} that extends the CL framework by using label information and extends the set of positive samples with samples of the same class. Let us briefly discuss the SCL framework that our work is related to. For each input sample $x$, we generate two random augmentations of it - $t_1(x), t_2(x)$, then both augmentation are separately input to the same base encoder $f(\cdot)$(ResNet\cite{he2016deep} backbone), resulting in a pair of representation vectors. The representation vectors are fed to a two-layer multi-layer perceptron (MLP) network $g(\cdot)$ that maps representations to the space where contrastive loss is applied, and the loss can be defined as:
\begin{equation} \label{eq:1}
     \mathcal{L}_{SC} = \\  \sum_{i\in I} \frac{-1}{|P(i)|} \sum_{p\in P(i)} \log \frac{\exp (z_i \cdot z_p/ \tau)}{\sum_{a\in A(i)} \exp (z_i \cdot z_a/ \tau)}
\end{equation}

\noindent Where, $i\in I \equiv \{1,...,2N\}$ is the index of an arbitrary augmented sample,  $z_\ell = g(f(t_j(x)))$, the $\cdot$ symbol denotes the inner (dot) product, $\tau \in \mathbb{R}^+$ is an scalar temperature parameter, $A(i) \equiv I\setminus i$, $P(i)\equiv \{p\in A(i) : y_p= y_i\}$ is the set of indices of all positives in the multiviewed batch distinct from $i$ and $|P(i)|$ is its cardinality. The definition of $P(i)$ spots the difference between CL and SCL, in SCL the set of positive samples is much larger and consists not only of sample with different augmentation, but all samples in the multiviewed batch with the same label.\\

\noindent\textbf{Adversarial training and self-supervised learning.} Most recent contrastive learning literature shows their use to improve natural accuracy. However, it has been shown that feature consistency w.r.t perturbation can improve adversarial robustness\cite{uesato2019labels}, and in this work we believe that feature consistency can further boost adversarial robustness, as adversarial examples might be the result of non-smooth feature space, i.e attacking a sample with small perturbation results in large feature change and misclassification. Few recent works explore the connection between SSL and adversarial training. \cite{chen2020adversarial} proposed to use several known self-supervised task losses like predicting rotation, permutation, and correct patches to pretrain a model and study the effect on its robustness. \cite{jiang2020robust,kim2020adversarial} used SSL to unsupervised learn robust feature representation and improve model robustness. Both methods proposed a new way to generate adversarial examples based on the CL loss instead of regular label-based losses(e.g., cross-entropy).\cite{wu2021two} proposed to train two encoders: clean trained and adversarially trained, and use two loss functions: contrastive loss to minimize feature inconsistency between natural and adversarial samples, and CE loss to promote high classification accuracy. The adversarial samples are used in the self-supervised feature representation training to achieve feature robustness. 

%% file: section/method.tex
\section{Contrastive Learning with Adversarial Features(CLAF)}
\label{sec:method}
In this section, we introduce CLAF, a supervised contrastive learning framework with robust feature representation. The method learns robust feature representations and achieves state-of-the-art robust accuracy while also achieving the highest clean accuracy compared to previous adversarial training-oriented methods. Our framework is composed of five main parts:
\begin{itemize}
    \item \emph{Data augmentation} module - Given an input sample $x$, it generates two random augmentations of it, each includes a small subset of the sample information and represents a different view of it. 
    
    \item Neural network \emph{encoder} $f(\cdot)$ that maps an input sample $x'$ to a feature representation vector $v = f(x')\in \mathbb{R}^d$.
    
    \item \emph{Projection head} - A small neural network $g(\cdot)$ that maps representation $v$ to the space where contrastive loss is applied $z = g(v)\in \mathbb{R}^p$.
    
    \item \emph{Classification head} - Single linear layer $c(\cdot)$ used to map the feature representation vector to a prediction vector $l = c(v) \in \mathbb{R}^n$, where $n$ is the number of classes.
    
    \item \emph{Attacker} module - Given a model, input sample, and true label, the attacker generates an adversarial sample similar to the original sample (under given distance metric) that causes misclassification. 
\end{itemize}

\begin{algorithm}[t]
  \caption{Pseudo-code of CLAF SCL with adversarial examples stage phase in PyTorch-like style.}
  \label{alg:code}
  \algcomment{\fontsize{7.2pt}{0em}\selectfont
  \vspace{-2.em}
  }
  \definecolor{codeblue}{rgb}{0.580,0.337,0.447}
  \lstset{
     backgroundcolor=\color{white},
     basicstyle=\fontsize{7.2pt}{7.2pt}\ttfamily\selectfont,
     columns=fullflexible,
     breaklines=true,
     captionpos=b,
     commentstyle=\fontsize{7.2pt}{7.2pt}\color{codeblue},
     keywordstyle=\fontsize{7.2pt}{7.2pt},
  %  frame=tb,
  }

\begin{lstlisting}[language=python, mathescape=true]
# g: projection head
# f: encoder
# c: classification head
# eps:  perturbation bound
# K: attack steps
# optimizer_classifier: optimizer for c parameters
# optimizer_features: optimizer for f and g parameters

# Linear classifier training
for _ in range(N):
    for x, y in loader: #  x: data, y: labels
        # generate adversarial examples with attacker
        x_adv = attack(f, c, x, y, K, eps)
        
        logits_adv = c.forward(f.forward(x_adv))
        # SGD update: linear classifier
        loss =  CrossEntropyLoss(logits_adv, y)
        loss.backward()
        optimizer_classifier.step()
        
# Encoder and projection head training    
for x, y in loader:#  x: data, y: labels
    x1 = aug(x) # a randomly augmented version
    x2 = aug(x) # another randomly augmented version
    # generate adversarial examples with attacker
    x_adv = attack(f, c, x, y, K, eps)
    
    # compute latent vectors
    f1 = g.forward(f.forward(x1))
    f2 = g.forward(f.forward(x2))
    f_adv = g.forward(f.forward(x_adv))
    features = torch.cat([f1, f2, f_adv], dim=1)
    
    # compute supervised contrastive loss
    loss = scl_loss(features, y)
    loss.backward()
    optimizer_features.step()
\end{lstlisting}
\end{algorithm}

\noindent The method training is divided into three stages:\\\\
\textbf{Supervised contrastive learning.} We start with vanilla supervised contrastive learning, same as \cite{khosla2020supervised}, for a relatively small number of epochs. Given the current batch of size $N$, for each sample in the batch, we generate two random augmentations of the sample, obtaining $2N$ augmented samples $\mathcal{D}_1=\{x_i\}_{i=1}^{i=2N}$. For each augmented sample $x_i\in\mathcal{D}_1$ we calculate the latent vector  $z_i=g(f(x_i)) \in \mathbb{R}^p$, where $f$ is the encoder and $g$ is the projection head. We update networks $f$ and $g$ to minimize $\mathcal{L}_{SC}$.\\\\
\textbf{Supervised contrastive learning with adversarial examples.} In this stage our goal is to add for each sample's positive set another positive example - an adversarial image. However, as discussed earlier, in order to create adversarial examples using known adversaries, e.g. PGD\cite{madry2017towards}, we need a classifier model. To overcome this hurdle, we train a linear classifier in parallel to the encoder training. We use a single linear layer that maps representation vector $v$ to a prediction vector $l \in \mathbb{R}^n$, where $n$ is the number of classes. At the beginning of each training epoch, we adversarially train, for a fixed and relatively small number of epochs, a linear classifier on top of the frozen encoder $f$. We perform this classifier retraining after every encoder parameter change in order to keep the linear classifier updated with the current feature representation learned by the encoder, and afterward train the encoder with reliable adversarial samples.
Later, in the attack algorithm, we use the encoder and the linear classifier as one pipeline to craft, for each sample, the adversarial example. Now, we get $2N$ augmented samples and another $N$ adversarial samples, obtaining in-total $3N$ samples used for training -- $\mathcal{D}_2\{x_i\}_{i=1}^{3N}$. The training continues the same way as in the regular SCL phase - for each sample $x_i \in \mathcal{D}_2$ we calculate it's latent vector $z_i$ and update network $f$ and $g$ to minimize $\mathcal{L}_{SC}$. The description of this step can be found in Algorithm  \ref{alg:code}.\\\\
\textbf{Linear evaluation.} To evaluate the quality of the learned visual representation, we leverage the widely-used linear evaluation criteria\cite{chen2020simple, zhang2016colorful}. We freeze the encoder $f$ weights and on top of it we train a linear layer $c(\cdot)$ using clean samples and labeled data. Since we keep the encoder untouched, this test can be seen as a proxy to the representation learned.

%% file: section/experiments.tex
\section{Experiments}
\label{sec:experiments}

We now present an empirical evaluation of the proposed method by measuring clean accuracy and robust accuracy and show a comparison of our method with previous adversarial learning methods. The code to reproduce the
experimental results is available at \url{https://github.com/AdirRahamim/CLAF}.
\subsection{Experimental settings}
We use CIFAR10\cite{krizhevsky2009learning} as the benchmark dataset in our experiment. It has 50k training images and 10k test images. For the encoder network $f(\cdot)$ we used ResNet-18 backbone\cite{he2016deep}. We take the output after the average pooling layer as the representation vector $v$, thus getting a vector of size 512. As the projection head, we used a MLP with a single hidden layer of size 512 and an output vector of size 128.  For encoder and projection head optimization we used SGD optimizer, with an initial learning rate of 0.05 with cosine learning rate decay. We experimented with a batch size of 256, and we ran the supervised contrastive learn phase for 60 epochs and supervised contrastive learn with adversarial examples phase for another 140 epochs(200 epochs in total).\\\\
For training of the classifier used to generate adversarial examples (denoted as $c$ in Algorithm \ref{alg:code}), we used a single linear layer of size 10. To generate adversarial examples during adversarial training of the classifier, we used the $\ell_\infty$ PGD attack\cite{madry2017towards} with hyperparameters $\epsilon=8/255$, $\eta=2/255$ and $k=5$, and we trained it each time for 5 epochs using Adam optimizer with a learning rate of 0.001. We used the cross-entropy(CE) loss for training. For the linear evaluation phase, we train a linear layer on top of the frozen encoder $f$. We train the linear layer for 100 epochs with Adam optimizer with an initial learning rate of 0.001 and cosine learning rate decay. We used the cross-entropy(CE) loss for training.
For robustness evaluation, we used the $\ell_\infty$ PGD attack\cite{madry2017towards} with hyperparameters $\eta=2/255$, $k=10$ and $\epsilon=8/255$ or $\epsilon=16/255$.

\begin{table}[htb]
\centering
\begin{tabular}{cccccc}
\hline
\multirow{2}{*}{\begin{tabular}[c]{@{}c@{}}Method\\ orientation\end{tabular}} & \multirow{2}{*}{Method} & Nat.          &     & \begin{tabular}[c]{@{}c@{}}PGD\\ (10)\end{tabular} & \begin{tabular}[c]{@{}c@{}}PGD\\ (10)\end{tabular} \\ \cline{3-6} 
                                                                              &                         &               & $\epsilon$ & 8/255                                              & 16/255                                             \\ \hline
\multirow{2}{*}{Nat.}                                                      & Natural                 & 95.5          &     & 0.0                                                & 0.0                                                \\
                                                                              & SCL\cite{khosla2020supervised}                     & 92.6          &     & 18.3                                               & 9.1                                                \\ \hline
\multirow{4}{*}{Adv.}                                                  & AT\cite{madry2017towards}                      & 84.8          &     & 44.7                                               & 25.9                                               \\
                                                                              & TRADES\cite{zhang2019theoretically}                  & 83.7        &     & 51.4                                               & 34.3                                               \\
                                                                              & RoCL-AT\cite{kim2020adversarial}                 & 80.2*          &     & 40.7*                                               & 22.8*                                               \\
                                                                              & Ours                    & \textbf{92.4} &     & \textbf{60.4}                                      & \textbf{48.3}                                      \\ \hline
\end{tabular}
\caption{Experimental results for clean and robust accuracy on CIFAR-10 dataset. All models are trained on ResNet-18. Natural denotes standard training with cross-entropy(CE) loss, AT denotes Madry adversarial training\cite{madry2017towards} and SCL is the supervised contrastive learning\cite{khosla2020supervised}. * is the reported results of \cite{kim2020adversarial}. $(\cdot)$ denotes the number of PGD steps.}
\label{tab:exp}
\end{table}

\begin{figure}[htb]
    \centering
    \includegraphics[width=1.0\linewidth]{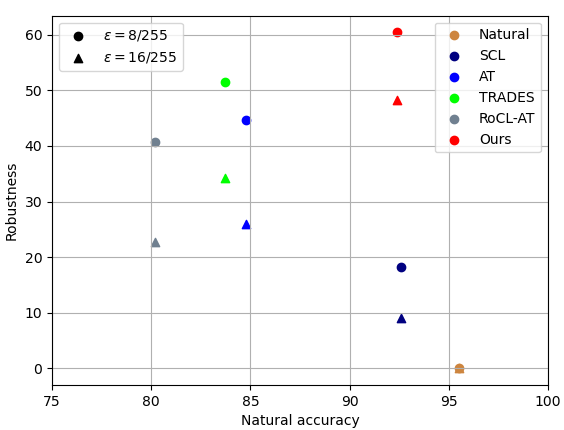}
    \caption{Natural accuracy and robust accuracy under $\ell_\infty$ PGD-10 attack with $\epsilon=8/255$ and $\epsilon=16/255$ of our method compared to other baselines.}
    \label{fig:accuracies}
\end{figure}

\subsection{Main results}
We report the clean and robust accuracy of previous state-of-the-art methods and our method in Table \ref{tab:exp}.
We achieve a new state-of-the-art robust accuracy against $\ell_\infty$ PGD-10 attack for both $\epsilon=8/255$ and $\epsilon=16/255$, significantly outperforms all recent baselines by a large margin, while keeping the highest clean accuracy(92.4\%) among all methods proposed to improve robust accuracy by adversarial training.
Compared to standard supervised contrastive learning\cite{khosla2020supervised}, our method significantly improves robust accuracy, which demonstrates that minimizing feature-space domain shift between natural and adversarial images indeed improves robust accuracy. Compared to other adversarially trained models, our method outperforms both AT\cite{madry2017towards} and TRADES\cite{zhang2019theoretically} methods. Moreover, our method outperforms the recent adversarial self-supervised contrastive learning method - RoCL-AT\cite{kim2020adversarial}, which adversarially pre-trains an encoder with adversarial examples founded by self-supervised contrastive loss and performs linear evaluation with supervised adversarial training.

%% file: section/conclusion.tex
\section{Conclusion}
\label{sec:CONCLUSION}

In this paper, we proposed a novel method for improving model robustness to adversarial attacks. We proposed a new idea of adding adversarial examples to sample positive set and adversarially train a linear classifier to generate them. We suggested the novel idea to retrain the linear classifier at each encoder train epoch to avoid feature inconsistency and keep linear classifier up-to-date with learned latent representation space. We demonstrated that adversarial robustness can be improved by maximizing the similarity between a transformed sample and generated adversarial sample of it, using the principle of supervised contrastive learning. We validated our method against the previous state-of-the-art methods and achieved superior clean and robust accuracy.

%% file: section/ablation.tex
\section{Ablation studies}
\label{sec:Ablation}

\begin{table}[tb]
\centering
\begin{tabular}{cccc}
\hline
\begin{tabular}[c]{@{}c@{}}Classifier train\\ type\end{tabular} & Nat. & \begin{tabular}[c]{@{}c@{}}PGD\\ 8/255\end{tabular} & \begin{tabular}[c]{@{}c@{}}PGD\\ 16/255\end{tabular} \\ \hline
Natural                                                         & 92.3 & 56.1                                                & 47.5                                                 \\
Adversarial                                                     & 92.4 & 60.4                                                & 48.3                                                 \\ \hline
\end{tabular}
\caption{Adversarial and natural trained linear classifier, denoted as $c$ in Algorithm \ref{alg:code}.}
\label{tab:adversrailVsNatural}
\end{table}

\begin{table}[tb]
\centering
\begin{tabular}{cccc}
\hline
PGD steps & 20    & 40    & 100   \\ \hline
Acc.      & 60.12 & 60.05 & 60.03 \\ \hline
\end{tabular}
\caption{Accuracy against $\ell_\infty$ PGD attack with different number of steps.}
\label{tab:pgdsteps}
\end{table}

\begin{table}[t!]
\centering
\begin{tabular}{cccc}
\hline
\begin{tabular}[c]{@{}c@{}}Linear evaluation\\ type\end{tabular} & Nat. & \begin{tabular}[c]{@{}c@{}}PGD\\ 8/255\end{tabular} & \begin{tabular}[c]{@{}c@{}}PGD\\ 16/255\end{tabular} \\ \hline
Natural                                                          & 92.4 & 60.4                                                & 48.3                                                 \\
Adversarial                                                      & 81.9 & 50.7                                                & 36.5                                                 \\ \hline
\end{tabular}
\caption{Adversarial and natural linear evaluation.}
\label{tab:advEvaluation}
\end{table}

\begin{table}[t!]
\centering
\begin{tabular}{cccc}
\hline
\begin{tabular}[c]{@{}c@{}}Linear\\ classifier\end{tabular}    & Nat. & \begin{tabular}[c]{@{}c@{}}PGD\\ 8/255\end{tabular} & \begin{tabular}[c]{@{}c@{}}PGD\\ 16/255\end{tabular} \\ \hline
Reinitialized & 92.2 & 60.8                                                & 49.5                                                 \\ 
Continuous    & 92.4 & 60.4                                                & 48.3                                                 \\ \hline
\end{tabular}
\caption{Reinitialize linear classifier $c$ before each training phase, and continue training from last state(continuous).}
\label{tab:reinitialized}
\end{table}

\begin{table}[t!]
\centering
\begin{tabular}{cccc}
\hline
Linear evaluation                                                               & Nat. & \begin{tabular}[c]{@{}c@{}}PGD\\ 8/255\end{tabular} & \begin{tabular}[c]{@{}c@{}}PGD\\ 16/255\end{tabular} \\ \hline
Reinitialized                                                                   & 92.4 & 60.4                                                & 48.3                                                 \\
\begin{tabular}[c]{@{}c@{}}Continue linear\\ classifier c training\end{tabular} & 92.4 & 60.9                                                & 48.3                                                 \\ \hline
\end{tabular}
\caption{Linear evaluation phase using reinitialized linear classifier and continue linear classifier $c$ in Algorithm \ref{alg:code} training.}
\label{tab:continueC}
\end{table}

\textbf{Adversarial linear evaluation.} In the linear evaluation phase, we freeze the encoder $f$ and on top of it we naturally train linear layer $c$. We can adversarially train the linear layer $c$ in the same adversarial training procedure. We adversarially train a linear classifier with adversarial images generated by $\ell_\infty$ PGD\cite{madry2017towards} attack with parameters $k=10$, $\epsilon=8/255$, $\eta=2/255$. The results in Table \ref{tab:advEvaluation} shows that naturally trained linear classifier in the linear evaluation phase performs better than adversarially trained linear classifier.\\

\noindent\textbf{Natural trained linear classifier.} In Algorithm \ref{alg:code}, the classifier $c$ is adversarially trained. However, it is possible to train it with natural examples. The comparative study in Table \ref{tab:adversrailVsNatural} shows that the adversarial training of the linear classifier $c$ improves both the natural and adversarial accuracy of the model.\\

\noindent\textbf{Use linear classifier $c$ in the linear evaluation phase.} In the linear evaluation step, we train a reinitialized linear classifier. We also experimented to use the linear classifier $c$ from Algorithm \ref{alg:code}, that we used to craft the adversarial examples, for the linear evaluation phase and train its last state for another 100 epochs. Table \ref{tab:continueC} shows that using linear classifier $c$ achieves similar results, with a small improvement of robustness under $\ell_\infty$ PGD attack with $\epsilon=16/255$.\\

\noindent\textbf{Reinitialize linear classifier $c$ in each epoch.} At the beginning of each encoder train epoch, we train the linear classifier $c$ for a small number of epochs, where we continue the training from the last state. We also examined reinitializing $c$ before each training phase. Table \ref{tab:reinitialized} shows that reinitializing the linear classifier improves robust accuracy against $\ell_\infty$ PGD attack - in 0.4\% for $\epsilon=8/255$ and for $\epsilon=16/255$ it improves in $1.2\%$, however it achieves $0.2\%$ lower accuracy on natural images.\\

\noindent\textbf{Robustness under more PGD attack steps.} We further validate the robustness of CLAF under various number of PGD attack iterations. The results on table \ref{tab:pgdsteps} shows that our method remains robust under higher number of steps (e.g., 60.03\% accuracy under 100 steps).